\pdfoutput=1

\documentclass[11pt]{article}

\usepackage{arabtex}
\usepackage[utf8]{inputenc}
\usepackage{utf8}
\setcode{utf8}

\usepackage[]{ACL2023}

\usepackage{multirow}
\usepackage{times}
\usepackage{booktabs}
\usepackage{amsfonts}
\usepackage{tabularray}

\PassOptionsToPackage{dvipsnames,table}{xcolor} 

\usepackage{comment}
\usepackage{array}      
\newcolumntype{L}[1]{>{\raggedright\arraybackslash}p{#1}}
\usepackage{color}
\usepackage{tabularx}  
\usepackage{booktabs}  
\usepackage{latexsym}

\newcommand{\ds}{\emph{LAILA}} 
\usepackage{paralist}
\usepackage{enumitem}
\usepackage[normalem]{ulem}

\usepackage[T1]{fontenc}

\usepackage[utf8]{inputenc}

\usepackage{microtype}

\usepackage{inconsolata}

\usepackage{todonotes} 

\definecolor{mod}{HTML}{bfd8bd}
\definecolor{subs}{HTML}{88d4ab}
\definecolor{perfect}{HTML}{469d89}
\usepackage{color, colortbl}

\newcommand{\highlight}[2]{\colorbox{#1}{#2}}

\title{LAILA: A Large Trait-Based Dataset for Arabic Automated Essay Scoring}

\author{
  \textbf{May Bashendy}$^1$, \textbf{Walid Massoud}$^1$, \textbf{Sohaila Eltanbouly}$^1$, \textbf{Salam Albatarni}$^1$ \\
  \textbf{Marwan Sayed}$^1$, \textbf{Abrar Abir}$^2$, \textbf{Houda Bouamor}$^2$, \textbf{Tamer Elsayed}$^1$ \\[1ex]
  $^1$Qatar University \\
  $^2$Carnegie Mellon University in Qatar \\[1ex]
  \texttt{\{ma1403845, wmassoud, se1403101, sa1800633, me2104862, telsayed\}@qu.edu.qa}, \\
  \texttt{\{aabir, hbouamor\}@cmu.edu}
}

\begin{document}
\maketitle
\begin{abstract}
        Automated Essay Scoring (AES) has gained increasing attention in recent years, yet research on Arabic AES remains limited due to the lack of publicly available datasets. To address this, we introduce \ds{}, the \emph{largest publicly available} Arabic AES dataset to date, comprising 7,859 essays annotated with \emph{holistic} and \emph{trait-specific} scores on \emph{seven} dimensions: relevance, organization, vocabulary, style, development, mechanics, and grammar. We detail the dataset design, collection, and annotations, and provide benchmark results using state-of-the-art Arabic and English models in prompt-specific and cross-prompt settings. \ds{} fills a critical need in Arabic AES research, supporting the development of robust scoring systems. 

\end{abstract}

\section{Introduction}


Automated Essay Scoring (AES) has emerged as a key area of research in writing assessment, as it not only minimizes the time and effort required for manual grading but also enables educators to provide timely and detailed feedback. By emphasizing linguistic quality rather than factual recall, AES systems assess writing in a manner that is independent of specific curricular content. 

Since the 1960s, most AES research has predominantly focused on English \cite{page1966imminence}, supported by the availability of several large-scale public datasets.  In contrast, Arabic AES research faces a significant data scarcity problem, as publicly available annotated datasets remain limited. Existing Arabic resources are often small in scale such as QAES \cite{bashendy-etal-2024-qaes}, lack trait-specific\footnote{A trait is an aspect of student writing, e.g., organization.} annotations as in ZAEBAC \cite{Habash2022}, or consist solely of unannotated essays such as ALC.\footnote{\url{https://www.arabiclearnercorpus.com}} 
These limitations have constrained the development of robust and scalable Arabic AES systems, despite some promising efforts \cite{Alghamdi2014, gaheenoptimized}.

\begin{table}
\centering
\small
\setlength{\tabcolsep}{2pt}
\begin{tabular}{ll}
\hline
\textbf{Trait} & \textbf{Description} \\
\hline
Relevance {\scriptsize \<الصلة بالموضوع>}
& Alignment with the prompt
\\

Organization {\scriptsize \<الهيكل العام>}
& The structure of the essay \\

Vocabulary {\scriptsize \<المفردات>}
& Word choice and variety
\\

Style {\scriptsize \<الأسلوب والتماسك البنائي>}
& Linking 
and transitions 
\\

Development {\scriptsize \<تطور الأفكار والمضمون>}
& Clarity and support of ideas
\\

Mechanics {\scriptsize \<الإملاء والترقيم والتنسيق>}
& Spelling and punctuation \\

Grammar {\scriptsize \<البناء والتراكيب>}
& Grammatical accuracy 
\\
\hline
Holistic {\scriptsize \<التقييم الكلي>}
& Overall writing quality \\
\hline
\end{tabular}
\caption{Brief description of the traits in \ds.}
\label{tab:traits_desc}
\end{table}

\begin{table*}[h]
\small
\centering
\begin{tblr}{
  width=0.9\textwidth,
  colsep=4pt,
  column{2} = {c},
  column{3} = {r},
  column{4} = {c},
  column{5} = {c},
  column{8} = {c},
  column{9} = {c},
  column{10} = {c},
  hline{1-2,9,15-16} = {-}{},
}
\textbf{Dataset} 						& \textbf{Lang} & \textbf{Essays} & \textbf{Prompts} & \textbf{Len} & \textbf{Level} & \textbf{L1/L2} & \textbf{HOL} & \textbf{Traits} & \textbf{Public}		\\
ASAP++	\cite{mathias2018asap++}		             				& EN            & 12,978          & 8                & 281          & G7-10          & L2              & \checkmark   & \checkmark      & \checkmark			\\

TOEFL11  	\cite{blanchard2013toefl11}		             				& EN            & 12,100         & 8                &     348      &   -       &        L2       &  \checkmark   & $\times$        & 	$\times$  
\\
ELLIPSE \cite{crossley2023english}          			& EN            & 6,500           & 44               & 427          & G8-12          & Both       & \checkmark   & \checkmark      & \checkmark	\\
PERSUADE \cite{crossley2023large}         			& EN            & 25,000          & 15               & 418          & G6-12          & L1          & \checkmark   & $\times$        & \checkmark	\\
ACEA \cite{he-etal-2022-automated}            			& ZH            & 1,220           & -                & -            & G12            & L1          & \checkmark     & \checkmark      & $\times$		\\
TCFLE-8 \cite{wilkens-etal-2023-tcfle}         			& FR            & 6,569           & 3                & 119          & -         & Both	      & \checkmark   & $\times$        & \checkmark	\\
MERLIN \cite{boyd2014merlin}          				& Eur    & 2,287           & -                & -            & -         & Both            & \checkmark   & $\times$        & \checkmark        \\
ZAEBUC \cite{Habash2022}          				& AR            & 214             & 3                & 156          & College      & L2              & \checkmark     & $\times$        & \checkmark                \\
QCAW \cite{Ahmed2024}            				& AR            & 195             & 2                & 499          & College       & L1          & \checkmark   & $\times$        & $\times$                \\
Abbir \cite{Alghamdi2014}           				& AR            & 640             & 2                & 150          & College       & L1& \checkmark & $\times$        & $\times$        \\
AAEE \cite{Azmi2019}            				& AR            & 350             & 8                & -            & G7-12          & L1          & \checkmark   & $\times$        & $\times$        \\
QAES \cite{bashendy-etal-2024-qaes}            			& AR            & 195             & 2                & 489          & College       & L1          & \checkmark   & \checkmark                & \checkmark                \\
TAQEEM 2025 \cite{taqeem2025}            		& AR            & 1,265             & 4                & 151          & G10-12       & L1          & \checkmark   & \checkmark                & \checkmark                \\
\textbf{LAILA}  						& AR            & 7,859           & 8                & 171            & G10-12         & L1          & \checkmark          & \checkmark             & \checkmark             
\end{tblr}
\caption{Comparison of \ds{} with existing essay datasets. ``Len'' is average essay length in words; ``HOL'' indicates holistic scoring; ``Eur'' covers German, Italian, and Czech; ``L1/L2'' denotes native or second-language learners; ``Public'' here means freely available.}
\label{tab:related_work}
\end{table*}

In this paper, we introduce {\ds},\footnote{\ds{} stands for "\textbf{L}inguistic \textbf{A}ssessment with tra\textbf{I}ts for \textbf{L}earning \textbf{A}rabic," and is pronounced in Arabic as {\small ``\<لَيْلَى>''}.} the \emph{first large-scale} Arabic AES dataset featuring both holistic and trait-specific annotations of seven writing proficiency traits, described in Table~\ref{tab:traits_desc}. The dataset comprises 7,859 essays across 8 distinct prompts making it the most extensive publicly available Arabic resource of its kind.\footnote{A prompt is the text that describes an essay writing task.} To construct {\ds}, our team of researchers conducted a large-scale data collection effort over the course of one academic year. Under Institutional Review Board (IRB) approval,\footnote{IRB Number: QU-IRB 159/2024-EA} we visited 24 high schools in Qatar and collected essays directly from 4,372 students in authentic classroom settings. We describe the complete dataset development pipeline, from data collection design to annotation procedures and public release, to facilitate future research on Arabic AES and writing assessment. Our contributions are as follows:

\begin{itemize}[leftmargin=1em, itemsep=-1pt, topsep=0pt]
    \item \textbf{Novel Dataset}: We introduce the first large-scale dataset of Arabic essays, filling a key gap in the existing Arabic AES literature.  
    \item \textbf{Annotation Guidelines}: We develop and share comprehensive guidelines for data annotation to ensure transparency and reproducibility.
    \item \textbf{Trait-Specific Annotations}: We provide detailed trait-specific annotations that capture multiple dimensions of writing proficiency.  
    \item \textbf{Benchmarking}: We establish baseline AES results for Arabic under two evaluation setups: prompt-specific\footnote{Trains and tests on essays from a single prompt.} and cross-prompt.\footnote{Trains on several prompts and tests on unseen ones.}
    \item \textbf{Public Release}: We publicly release \ds{},\footnote{\url{https://gitlab.com/bigirqu/laila}} including essays with holistic and trait-specific annotations. We also release the benchmarking 
    code \footnote{\url{https://gitlab.com/bigirqu/laila-baselines}} to support replication and 
    future research.  
    
\end{itemize}

The remainder of this paper is organized as follows. Section~\ref{sec:related-work} reviews related work and existing AES datasets. Section~\ref{sec:design} outlines the data collection design principles, while Section~\ref{sec:dataset-construction} describes the dataset construction process, including school and prompt selection, essay collection, and annotations. Section~\ref{sec:dataset-analysis} presents dataset analysis, and Section~\ref{sec:dataset-release} details the public dataset release. Section~\ref{sec:exp} reports preliminary experiments and results, and Section~\ref{sec:conc} concludes with final remarks.
\section{Related Work}
\label{sec:related-work}
To contextualize our contribution, we review prior AES research across languages and modeling paradigms, outlining existing datasets and the evolution of Arabic AES from feature-based to neural and LLM-based approaches. Table~\ref{tab:related_work} summarizes major AES datasets across languages, underscoring the novelty and significance of \ds{} as a benchmark resource for future Arabic AES research.

\subsection{AES Datasets}

\paragraph{Non-Arabic AES Datasets}
Research on AES has advanced considerably in English, supported by large-scale, publicly available
datasets such as ASAP/ASAP++~\cite{mathias2018asap++}, TOEFL11~\cite{blanchard2013toefl11}, ELLIPSE~\cite{crossley2023english}, and PERSUADE~\cite{crossley2023large}. These resources have enabled benchmarking and model development for both prompt-specific and cross-prompt AES tasks. In contrast, other languages have far fewer resources. Some smaller datasets exist, including ACEA (Chinese; annotated but not public) \cite{he-etal-2022-automated}, TCFLE-8 (French; public but with only holistic scores) \cite{wilkens-etal-2023-tcfle}, and MERLIN (German, Italian, and Czech; public but limited to holistic annotations) \cite{boyd2014merlin}. While useful for AES studies, their limited scale and annotation depth restrict their utility.

\paragraph{Arabic AES Datasets}
Compared with English and other 
languages, Arabic AES lacks robust datasets. Few datasets have been developed, both public and proprietary, yet all face limitations in scale, annotation depth, or accessibility. Publicly available
resources include the Zayed Arabic English Bilingual Undergraduate Corpus (ZAEBUC) \cite{Habash2022}, which offers linguistic annotations (POS tagging, lemmatization, and spelling corrections) and CEFR proficiency levels but lacks trait-level AES labels. The Qatari Corpus of Argumentative Writing (QCAW) \cite{Ahmed2024} contains 195 essays with POS annotations but restricted holistic scores, while its extension, QAES \cite{bashendy-etal-2024-qaes}, introduces both holistic and trait-level annotations. Building on these efforts, ZaQQ \cite{elsayed2025zaqq} merges QAES, ZAEBUC, and QALB \cite{mohit-etal-2014-first}, annotated using both human raters and LLMs. 
More recently, the TAQEEM 2025 shared task dataset \cite{taqeem2025} released 1,265 essays across 4 prompts, and TAQAE \cite{sayed2025feature} combined QAES with 2 prompts from TAQEEM. It is important to note that \ds{} reuses the 2 prompt texts from QAES dataset (with no essays taken from it), and fully include TAQEEM 2025 dataset (4 prompts with 1,265 essays).

Beyond these datasets, several proprietary resources have been used, 
including Abbir \cite{Alghamdi2014}, which contains essays by Saudi university students with holistic scores, and AAEE \cite{Azmi2019}, comprising essays from grades 7--12 evaluated for content, style, and spelling. However, their unavailability limits reproducibility and progress in Arabic AES. 

Overall, existing Arabic AES resources remain fragmented and small in scale (typically under 2,000 essays), limiting robust model training and standardized benchmarking. 
In contrast to prior work, \textbf{\ds{}} provides a large-scale, publicly-available Arabic AES dataset with holistic and trait-specific annotations. In terms of scale and annotation richness, \ds{} is comparable to leading datasets in other languages and fills a critical gap in the current Arabic AES landscape. 

\subsection{Arabic AES Systems}
Despite the scarcity of datasets, several studies have explored Arabic AES. Early approaches primarily relied on handcrafted feature engineering \cite{alqahtani-alsaif-2020-automated,alsanie2022automatic}, and text similarity techniques \cite{abdeljaber2021automatic,9498119,al2019automated,alobed2021automated,Azmi2019}. More recent work has shifted towards leveraging pretrained language models, notably through fine-tuning AraBERT \cite{ghazawi2024automated}, integrating handcrafted features \cite{machhout2024enhanced}, or applying parameter-efficient adaptation methods \cite{mahmoud2024automatic}. Most recently, \citet{sayed2025feature} initiated cross-prompt research in Arabic AES by developing a set of engineered features and evaluating their effectiveness across feature-based and encoder-based models. Additionally, \citet{ghazawi2025well} pioneered the use of LLMs for Arabic AES, assessing systems such as GPT and LLaMA under various prompting configurations.  In this work, we benchmark \ds{} dataset with state-of-the-art (SOTA) Arabic and English models to establish strong and reproducible baselines for future research.

\begin{figure*}[t!]
    \centering
    \includegraphics[width=1\textwidth]
    {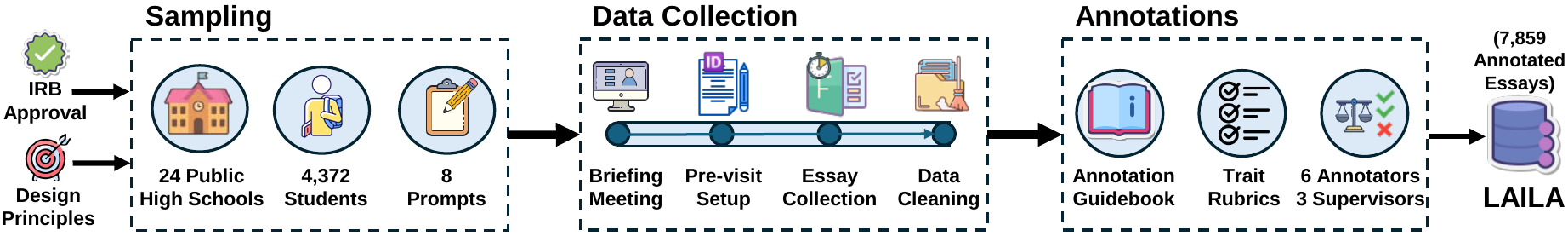}
    \caption{An overview of the construction process of \ds{} dataset.}
    \label{fig:laila_framework}
\end{figure*}

\section{Dataset Design Principles}
\label{sec:design}
The development of \ds{} was guided by nine core design principles [D1–D9], each reflecting deliberate choices to ensure a high-quality, representative, and ethically sound resource for Arabic AES. Below, we outline these principles. 

\begin{itemize}[leftmargin=0.8em, itemsep=-1pt, topsep=0pt]

\item \textbf{[D1] Ensuring Data Integrity:}
We designed \ds{} to capture authentic, classroom-produced writing created without external assistance. This design choice preserves genuine linguistic variation, 
learner errors, and authentic complexities essential for real-world AES applications.

\item \textbf{[D2] Maximizing Diversity:
}
To reflect the breadth of Arabic writing proficiency, \ds{} was planned to collect 
essays from multiple educational institutions and student populations with diverse academic and demographic backgrounds. Prompts were intentionally
varied across genres, ensuring cultural relevance and age appropriateness, thereby promoting model generalization and reducing bias.

\item \textbf{[D3] Achieving Gender Balance: 
}
We intentionally aimed to balance 
the number of essays from male and female students to promote fairness, reduce gender-based bias, support inclusive evaluation, and enhance model generalization. 

\item \textbf{[D4] 
Supporting
Full Trait Coverage:
}
\ds{} was built to capture multiple dimensions of writing quality through annotations of relevance, organization, vocabulary, style, development, mechanics, and grammar. This design supports 
fine-grained assessment and 
meaningful feedback.

\item \textbf{[D5] Ensuring 
Authentic Writing Conditions:}
To emulate real academic or high-stakes testing scenarios, \ds{} was designed to collect essays written under timed conditions using a digital submission platform.
This setup ensures consistency in data collection and eliminates manual transcription errors.

\item \textbf{[D6] Protecting Participant Privacy:}
To ensure ethical research and data protection, 
\ds{} was planned with strict privacy measures 
by anonymizing all essays and collecting them 
with informed consent from students and guardians. 

\item \textbf{[D7] Standardizing the Annotation Process:}
To ensure reliability, 
\ds{} aimed to use a standardized rubric for evaluating essays across all traits, with planned training for annotators to maintain consistent interpretations of the guidelines and achieve high inter-annotator agreement.

\item \textbf{[D8] Promoting 
Scalability and Reusability:}
\ds{} was planned at a large scale to enable robust AES model training. All data and annotations are provided in standardized, machine-readable formats with comprehensive documentation and an open license to support future reuse. 

\item \textbf{[D9] Supporting Multiple Evaluation Setups:}
We structured \ds{} to support both \textit{prompt-specific} and \textit{cross-prompt} AES setups, facilitating research on model generalization. 

\end{itemize}


\section{Dataset Construction}
\label{sec:dataset-construction}


The construction of \ds{} follows three stages: \emph{Sampling}, involving the selection of students, schools, and writing prompts; \emph{Data Collection}, covering essay collection and data cleaning; and \emph{Annotations}, focusing on scoring the collected essays. 
Figure~\ref{fig:laila_framework} shows an overview of the overall process.

\subsection{Sampling} 

The first stage of constructing \ds{} involved sampling schools, students, and writing prompts. Data were collected from 24 high schools in Qatar, evenly divided between female and male institutions [D3], selected for having Arabic as the primary language of their studies. 
The selected schools provided a representative sample encompassing diversity in geographic location, size, and academic performance [D2]. All participating institutions were public and followed a unified national curriculum, ensuring a consistent academic context for the writing tasks.

Students were drawn from grades 10 to 12 to capture a wide range of writing proficiency. This age group was chosen because students at this level typically produce essays of sufficient length and complexity to address a variety of prompts. To reflect the diversity of academic performance, approximately 30\% of participants were low-performing, 40\% medium-performing, and 30\% high-performing [D2]. Participation was voluntary and required 
parental consent to ensure compliance with ethical research standards [D6].

Writing prompts were selected to elicit critical thinking and showcase different aspects of writing ability. The dataset includes eight distinct prompts, five persuasive and three explanatory, designed to capture both persuasive reasoning and expository clarity [D2]. This diversity supports the development 
of AES models capable of generalizing across writing genres and skill levels.

\subsection{Data Collection}
The second stage of constructing \ds{} focused on the systematic collection of high-quality student essays under controlled 
conditions. This phase involved coordinating with participating schools, preparing the technical setup for essay submission, conducting supervised on-site writing sessions, and performing 
post-collection data cleaning to ensure completeness, authenticity, and reliability.

\definecolor{verylightgray}{gray}{0.98} 

\begin{figure*}[t]
\centering
\small
\fcolorbox{black}{verylightgray}{
\begin{minipage}{\textwidth}
\tiny
\color{brown}
\begin{arabtex}

\textbf{نص الموضوع:} 
يلجأ كثير من الشباب إلى السهر لساعات طويلة ليلًا، ما يؤثر في نشاطهم اليومي وأدائهم الدراسي.
اكتب مقالًا مكوَّنًا من ثلاثمائة (300) كلمة توضح فيه العوامل التي تدفع الشباب إلى السهر، وما يترتب على ذلك من آثار جسدية ونفسية، مُراعيًا سمات المقال التفسيري، وسلامة اللغة، ومُوظفًا علامات الترقيم وأدوات الربط بشكلٍ صحيح.
\end{arabtex}
\vspace{0.5em}
\color{brown}
\textbf{Prompt (English translation):} Many young people stay awake late at night, which affects their daily activity and academic performance. Write an essay of approximately 300 words explaining the factors that drive young people to stay up late, and the resulting physical and psychological effects, taking into account the characteristics of explanatory essay features, using correct language, punctuation, and linking words.

\vspace{0.5em}
\tiny
\color{darkblue}
\begin{arabtex}
\vspace{0.5em}

\textbf{الموضوع:}
لا شك ان ظاهرة السهر عند شباب هذا الجيل مشكلة تؤثر على نشاطهم الصحي و أدائهم الدراسي بشكل سلبي و غير صحي, فقد اثبتت الدراسات ان السهر لاوقاتٍ طويلةٍ للشباب و الاطفال مشكلة تؤثر على صحتهم النفسية و العقلية مما قد يؤدي الى تراجع مستواهم الدراسي و الصحي و تودي الى تقلبات بالمزاج عندهم, و يوجد عدة اسبابٍ و عوامل تدفع الشباب الى السهر لاوقاتٍ متأخرة و منها:

اولاً: التركيز في العاب الفيديو: فشباب هذا الجيل يقضون معظم اوقاتهم في اللهو و اللعب خاصة بالعاب الفيديو, فان العاب الفيديو مستويلةٌ على عقل شباب هذا الجيل مما جعلهم ينسون العالم الخارجي و يلتهون عن اعملاهم و ينسون الوقت الذي يقضونه على العاب الفيديو مما يؤدي الى تضييع تركيزهم و تلاف عقلهم نتيجة السهر لاوقاتٍ طويلة.

ثانياً: شرب المنبهات: فكثيرةً ما نرى شباباً باول اعمارهم يشربون المنبهات بشكلٍ مبالغٍ فيه لكي يستطيعُ السهر و اضاعة الوقت, فقد اثبتت الدراسات ان شرب المنبهات من عمر صغيرٍ قد يؤدي الى 
مشاكل بالقلب او بالصحة و بعض الاحيان تفعل مشاكل اكبر من هذه بكثير.

ثالثاً: استخدام الهواتف المحمولةؤ بشكلٍ مبالغٍ فيه: فقد اصبح شباب هذا الجيل يمتلكون الهواتف النقالة باحدث انواعها و ينسون الوقت بسببها فلا يفرقوا بين الليل و النهار لانهم ...
مشغولون بالهواتف و قد نسوا وقتهم و اعمالهم, و ايضا و قد ذكر في القرآن دليل على مخاطر السهر في قوله تعالى:(( و جعلنا الليل لبسا و جعلنا النهار معاشا )).

و ختاماً ارى ان السهر احد اسوء و اشد المشاكل التي قد يمر بها الشباب فهي تدمر العقل و  المستقبل و عواملها قد تؤدي احيانا الي الموت, ان السهر لاوقاتٍ طويلة مرض لعينٌ من الصعب التخلص منه. و انا ارى ان السهر احد
 احقر و اسوء المشاكل التي قد تدمر جيلنا الحالي او الاجيال القادمة و يجب على الجميع ان يتحد بحل الموضوع لانه قد يؤدي الى هلاك المجتمعات.

\vspace{0.5em}
\end{arabtex}
\tiny
\color{purple}
\begin{arabtex}
    
\textbf{الدرجات:} الصلة بالموضوع: <2>، الهيكل العام: <5>، المفردات: <4>، الأسلوب والتماسك البنائي: <5>، تطور الأفكار والمضمون: <5>، الإملاء والترقيم والتنسيق: <4>، البناء والتراكيب: <4>، التقييم الكلي: <29>

\end{arabtex}
\vspace{0.5em}
\color{purple}
\textbf{Scores (English translation):} Relevance: 2, Organization: 5, Vocabulary: 4, Style: 5, Development: 5, Mechanics: 4, Grammar: 4, Holistic: 29

\end{minipage}%
}
\caption{An example from \ds, containing the \color{brown} prompt (P7) \color{black} with its English translation, \color{black} \color{darkblue} essay (070773) \color{black} without translation, to preserve Arabic-specific linguistic errors, \color{black} and \color{purple} annotations (scores) \color{black}with their English translation.}
\label{fig:essay_example}
\end{figure*}

\paragraph{Briefing}To ensure a structured 
data collection, an online briefing was held with the heads of Arabic departments from all participating schools. It outlined the essay collection setup and emphasized two key requirements: the writing task had to be conducted in an exam-like setting without access to external materials [D1], and student participation requires \emph{both} parental and student consents.

\paragraph{Pre-visit Setup} 

To facilitate essay collection, we prepared printed sheets, one for each student, containing one persuasive and one explanatory prompt, along with a unique, pre-generated ID to maintain students' anonymity while enabling metadata tracking [D6]. We also configured a Microsoft Form for digital essay submission, set with a 65-minute time limit to enforce a time-restricted setup [D5].

\paragraph{On-site Visits}
Data collection spanned one academic year through 27 school visits (24 initial, three follow-ups), by nine team members, to meet the target student count. Each visit, which lasted around six hours, was supported by at least two team members, an IT teacher, and proctoring teachers. During each visit, students received the printed sheets and accessed the online submission form to submit their essays within the allotted time.


\paragraph{Data Cleaning}
The collected essays then underwent a cleaning process to ensure data quality before the annotation phase. First, we resolved duplicate and misidentified submissions, by correcting ID mismatches through manual verification and name cross-referencing. Next, a manual review eliminated submissions lacking meaningful content, e.g., irrelevant personal content. Lastly, essays with 10 or fewer words were excluded for being insufficient for AES analysis. This approach preserved the data integrity for subsequent processing.

\subsection{Data Annotation}
\label{sec:annotations}
The third stage of constructing \ds{} involved annotating the collected essays 
through hiring reliable annotators, adopting standardized rubrics, and developing an annotation guidebook. All annotations were managed and executed using an online platform. Figure \ref{fig:essay_example} shows a sample annotated essay.

\paragraph{Annotators}
The hired annotation team consisted of 6 annotators and 3 supervisors, all of whom were Arabic language teachers or lecturers with educational backgrounds. Five members of the team hold advanced degrees (MSc or PhD) in the Arabic language. The supervisors oversaw training, quality assurance, and dispute resolution, while the annotators performed the primary scoring tasks.

\paragraph{Annotation Guidelines}
All essays in \ds{} dataset were scored using the Core Academic Skills Test (CAST) rubric [D7], developed by the Qatar University Testing Center (QUTC).\footnote{\url{https://www.qu.edu.qa/sites/en_US/testing-center/TestDevelopment/cast}} The rubric covers 7 writing traits [D4]: Relevance (REL), Organization (ORG), Vocabulary (VOC), Style (STY), Development (DEV), Mechanics (MEC), and Grammar (GRA). Additionally, a Holistic score (HOL) was computed as the sum of the 7 trait scores. Six traits (all but REL) were rated on a 6-point scale (0 = lowest, 5 = highest), while REL was rated on a 3-point scale (0 = not relevant, 1 = partially relevant, and 2 = fully relevant). Furthermore, if an essay received a REL score of 0, all other trait scores were set to 0, as irrelevant responses are not subject to further evaluation. This mixed-scale 
allowed the rubric to capture both broad dimensions of writing proficiency and the 
degree to which essays addressed the assigned topic.

To ensure consistency, 2 supervisors developed an annotation guidebook\footnote{\url{https://gitlab.com/bigirqu/laila/-/raw/main/rubrics/annotation_guidebook.pdf}} with detailed terminology, exemplars, and annotated practices for each prompt type. Annotators were required to review these materials and complete training sessions before formal annotation. Moderation sessions followed, where discrepancies were discussed and interpretations of the rubric were harmonized.



\begin{table}[t]
\centering
\small
\setlength{\tabcolsep}{2.5pt}
\begin{tabular}{c l l r c r}
\hline
\textbf{P\#} & \textbf{Type} & \textbf{Topic} & \textbf{Essays} & \textbf{Len (Max)} & \textbf{A3} \\
\hline
P1 & EXP & Sports & 1,122 & 162 (696) & 10.4\% \\
P2 & PER & Social Media & 1,168 & 175 (690) & 15.5\% \\
P3 & PER & Technology & 521 & 159 (643) & 9.4\% \\
P4 & PER & Communication & 500 & 152 (682) & 15.0\% \\
P5 & EXP & Heritage & 1,181 & 157 (690) & 23.3\% \\
P6 & PER & Homework Load& 1,162 & 160 (706) & 20.3\% \\
P7 & EXP & Staying Up Late & 1,143 & 202 (689) & 10.6\% \\
P8 & PER & Video Games & 1,062 & 186 (690) & 15.7\% \\
\hline
\textbf{Total} &  &  & \textbf{7,859} &  &  \\
\hline
\end{tabular}%
\caption{Descriptive statistics of \ds{}. ``EXP'' and ``PER'' denote explanatory and persuasive prompts; ``Len'' is the average word count; ``A3'' indicates the percentage of essays annotated by a third annotator (supervisor).}
\label{tab:dataset_statistics}
\end{table}

\paragraph{Annotation Process}
The scoring process was conducted using the Assessment Gourmet Platform, which supports large-scale, anonymized essay scoring.\footnote{\url{https://g-assess.com}} The platform ensured that annotators were blinded to student identity [D6], while only supervisors had access to annotator metadata for monitoring purposes.

The essays were randomly distributed among the annotators to minimize systematic 
bias, and the scoring sessions were capped to prevent annotators' fatigue. 
Each essay was independently scored by two annotators across all traits. If the difference in HOL scores between the two annotators was less than 6 points,\footnote{The 6-point threshold was defined by QUTC experts, corresponding to an average of 1-point discrepancy across the 6 core traits, excluding REL, composing the holistic score.} the mean of the two scores was computed and then rounded down to the nearest integer; this rounded value was adopted as the final score for each trait. However, essays with large discrepancies between annotators ($\geq 6$ points difference in the HOL score) were flagged and escalated to a supervisor (a third annotator in this case), who provided the final adjudicated score and offered feedback to the annotators to improve alignment and consistency in subsequent batches of essays. This multi-layered workflow, calibration, double-scoring, adjudication, and feedback ensured reliable 
and replicable annotations. 


\section{Dataset Analysis}
\label{sec:dataset-analysis}

\ds{} comprises 7,859 essays collected across 8 distinct prompts: 3 explanatory and 5 persuasive (3,446 and 4,413 essays, respectively) [D8]. 
Table \ref{tab:dataset_statistics} shows that the number of essays per prompt ranges from 500 to 1,181. Essay lengths 
demonstrate consistency across prompts, with an average of 171 words and a maximum of 706 words.

\begin{figure}[t]
    \centering
    \includegraphics[width=\columnwidth]{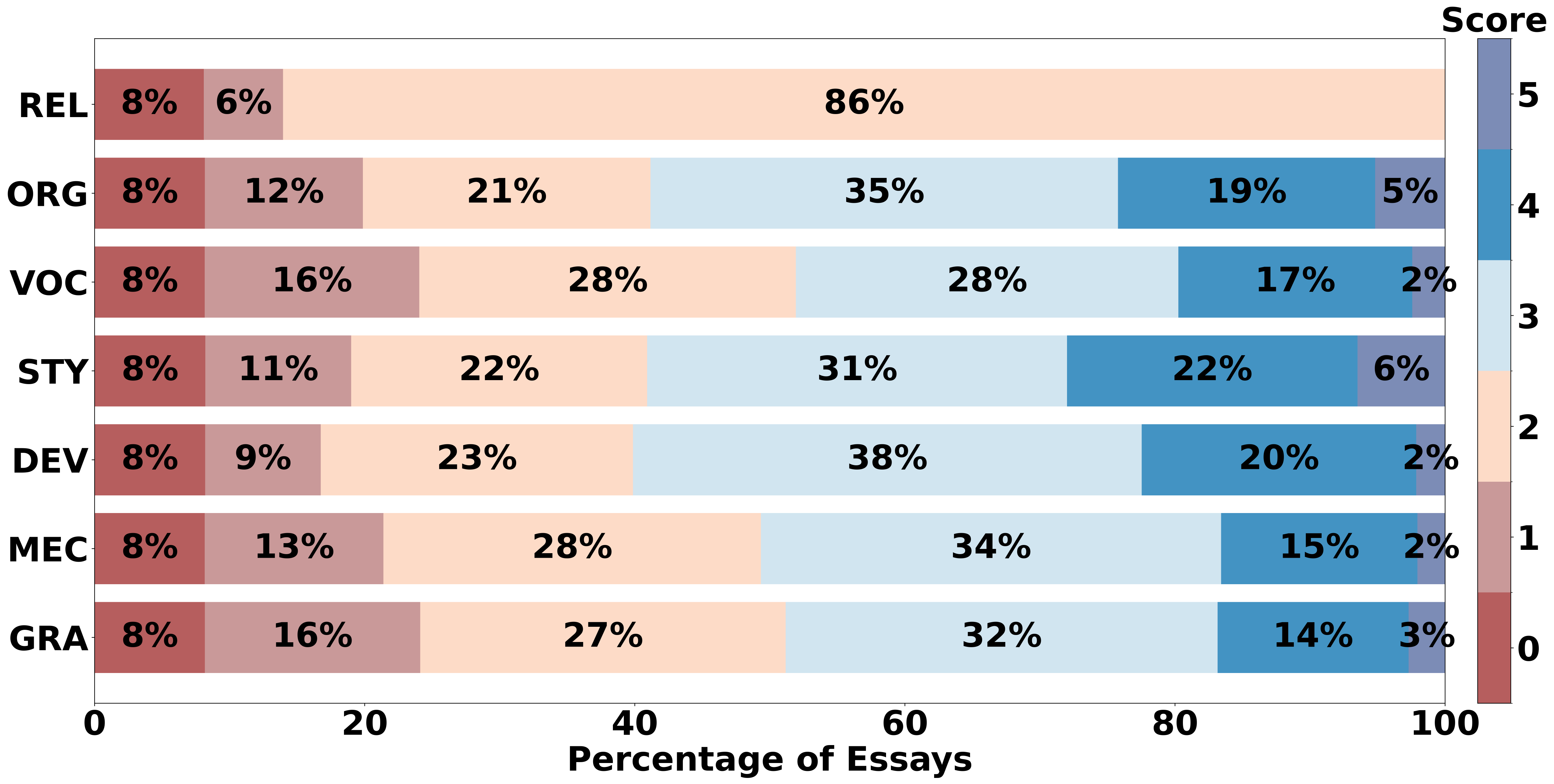}
    \caption{Overall score distributions \textit{per trait} in \ds.}
    \label{fig:dist_across_all_prompts}
\end{figure}

\paragraph{Score Distribution}
Figure \ref{fig:dist_across_all_prompts} shows the distribution of the trait scores on all prompts. Most traits follow a near-normal pattern on the 6-point scale, indicating overall positive essay quality. The REL scores show that 86\% of the essays agree strongly with the prompt. In contrast, GRA and VOC show the highest percentage of score 1 at 16\%, followed by MEC at 13\%, highlighting that these traits pose the greatest challenges. Other traits show a more balanced spread across mid-to-high scores, reflecting better performance. Extreme scores (0 or 5) are rare, with STY having the highest share of score 5 at 6\%. More details are provided in Appendix \ref{sec:app_dataset_analysis}.


\paragraph{Inter-Annotator Agreement (IAA)}
We compute IAA using Quadratic Weighted Kappa (QWK) \cite{QWK-IAA} to assess agreement 
between the two main annotators (A1 and A2), prior to adjudication. Table \ref{tab:IAA} reports QWK 
per trait and prompt, with agreement strength following \citet{IAA-scale}. Overall, IAA was \emph{substantial} across prompts, with an average QWK ranging from 0.66 (P5) to 0.75 (P1), showing strong consistency. 
However, variability emerges by prompt and trait. In particular, P5 showed the lowest average agreement, noting that it has the highest percentage of essays that require a third annotator (A3: 23.3\%; Table \ref{tab:dataset_statistics}). This suggests greater initial disagreement between A1 and A2 for P5, potentially due to less precise prompt wording that made consistent interpretation more challenging. Importantly, this does not reflect lower final label quality as such disagreements were flagged and resolved through adjudication by A3, who provided the final scores. 
Across traits, ORG showed the strongest agreement due to its structural clarity, whereas REL, DEV, and MEC had lower agreement, reflecting greater subjectivity in semantic traits.


\begin{table}[tb!]
\centering
\small
\setlength{\tabcolsep}{4.5pt}
\begin{tabular}{
    p{0.08\columnwidth}
    >{\centering\arraybackslash}p{0.07\columnwidth}
    >{\centering\arraybackslash}p{0.07\columnwidth}
    >{\centering\arraybackslash}p{0.07\columnwidth}
    >{\centering\arraybackslash}p{0.07\columnwidth}
    >{\centering\arraybackslash}p{0.07\columnwidth}
    >{\centering\arraybackslash}p{0.07\columnwidth}
    >{\centering\arraybackslash}p{0.07\columnwidth}
    >{\centering\arraybackslash}p{0.07\columnwidth}
}\hline
\textbf{Trait} & \textbf{P1} & \textbf{P2} & \textbf{P3} & \textbf{P4} & \textbf{P5} & \textbf{P6} & \textbf{P7} & \textbf{P8} \\ \hline
REL & \highlight{subs}{0.79} & \highlight{mod}{0.60} & \highlight{subs}{0.67} & \highlight{mod}{0.59} & \highlight{mod}{0.58} & \highlight{subs}{0.68} & \highlight{subs}{0.75} & \highlight{subs}{0.77} \\
ORG & \highlight{subs}{0.78} & \highlight{subs}{0.72} & \highlight{perfect}{0.83} & \highlight{subs}{0.77} & \highlight{subs}{0.74} & \highlight{subs}{0.78} & \highlight{subs}{0.77} & \highlight{subs}{0.78} \\
VOC & \highlight{subs}{0.74} & \highlight{subs}{0.71} & \highlight{subs}{0.76} & \highlight{subs}{0.69} & \highlight{subs}{0.71} & \highlight{subs}{0.75} & \highlight{subs}{0.79} & \highlight{subs}{0.80} \\
STY & \highlight{subs}{0.74} & \highlight{subs}{0.71} & \highlight{subs}{0.76} & \highlight{subs}{0.72} & \highlight{mod}{0.60} & \highlight{subs}{0.71} & \highlight{subs}{0.68} & \highlight{subs}{0.72} \\
DEV & \highlight{subs}{0.79} & \highlight{subs}{0.64} & \highlight{mod}{0.58} & \highlight{subs}{0.66} & \highlight{subs}{0.70} & \highlight{subs}{0.72} & \highlight{subs}{0.71} & \highlight{mod}{0.55} \\
MEC & \highlight{subs}{0.69} & \highlight{subs}{0.65} & \highlight{subs}{0.77} & \highlight{subs}{0.72} & \highlight{mod}{0.58} & \highlight{subs}{0.65} & \highlight{subs}{0.69} & \highlight{subs}{0.70} \\
GRA & \highlight{subs}{0.71} & \highlight{subs}{0.65} & \highlight{subs}{0.77} & \highlight{subs}{0.70} & \highlight{subs}{0.70} & \highlight{subs}{0.73} & \highlight{subs}{0.67} & \highlight{subs}{0.61} \\

\hline
AVG & \highlight{subs}{0.75} & \highlight{subs}{0.67} & \highlight{subs}{0.73} & \highlight{subs}{0.69} & \highlight{subs}{0.66} & \highlight{subs}{0.72} & \highlight{subs}{0.72} & \highlight{subs}{0.70} \\ \hline

\end{tabular}

\caption{Inter-Annotator Agreement assessed via QWK across all prompts. Colors indicate strength of agreement as \protect\highlight{mod}{Moderate}, \protect\highlight{subs}{Substantial}, and \protect\highlight{perfect}{Almost Perfect}.}
\label{tab:IAA}
\end{table}

\section{Dataset Release}
\label{sec:dataset-release}
We release \ds{} as an open-access resource for Arabic AES, including writing prompts, essays, scoring rubrics, annotation guidebook, and annotations [D8]. The annotations include scores assigned by the main annotators, the third annotator used for adjudication, and the final adjudicated scores covering both holistic and trait-specific dimensions. Moreover, we provide two predefined data splits aligned with the commonly-applied AES setups: \emph{prompt-specific} and \emph{cross-prompt} [D9]. For the prompt-specific setup, essays of each prompt were randomly split over 5 folds, to support a 5-fold cross-validation setting. For the cross-prompt setup, we support 8-fold leave-one-prompt-out cross-validation setting as follows. For each of the 8 prompts, the target prompt is held out entirely for testing, 2 prompts (1 persuasive and 1 explanatory, selected at random) are designated for development, and the remaining 5 prompts are designated for training. For reproducibility and fair of comparisons, splits are made publicly available.\footnote{\url{https://gitlab.com/bigirqu/laila}}

\section{Benchmarking Experiments}
\label{sec:exp}

To establish performance references, we benchmarked \ds{} under both prompt-specific and cross-prompt AES settings using strong baselines. For each setup, we investigate two research questions in the context of Arabic AES: \textbf{(RQ1)} how do different categories of models compare in overall performance?, and \textbf{(RQ2)} which models achieve the best results across individual traits?.

\subsection{Experimental Setup} 
\label{sec:experimental_setup}

\paragraph{Model Selection}
We selected a diverse set of models with varying architectures to establish strong baselines for \ds. The selection criteria included code availability, ease of implementation, and coverage of SOTA Arabic and English models. The models fall into three categories: feature-based (FB), encoder-based, and large language models (LLMs). LLMs were evaluated in zero-shot and few-shot settings, and all models followed a multi-task setup predicting holistic and trait scores jointly. Further details are provided in Appendix \ref{appendix:baselines}.
 
\paragraph{Prompt-specific Models}
\label{sec:prompt-sp}
We selected 4 FB models: Linear Regression (LR), Random Forest (RF), Extreme Gradient Boosting (XGB), and a feedforward Neural Network (NN). For encoder-based models, we fine-tuned two pre-trained SOTA systems: AraBERT~\cite{antoun2020arabert}, combined with handcrafted features \cite{sayed2025feature}, and AraT5 \cite{nagoudi2021arat5}, inspired by the strong performance of T5 in English AES \cite{do-etal-2024-autoregressive}.\footnote{\url{https://huggingface.co/aubmindlab/bert-base-arabert}, \url{https://huggingface.co/UBC-NLP/AraT5v2-base-1024}} For LLMs, we evaluated three Arabic-centric models: ALLaM \cite{bari2025allam}, Command-R7B-Arabic (R7B) \cite{alnumay2025command}, and Fanar \cite{fanarllm2025}, under zero-shot and few-shot (5-shot) settings. For the prompt-specific experiments, we used 5-fold cross-validation using the predefined 5 splits. In each cross-validation iteration, one fold (20\%) was used as the test set, and the remaining four folds were further split into training (70\%) and development (10\%) sets, with stratification applied to maintain consistent prompt distributions and ensure consistent evaluation across models.


 \begin{table*}[h]
\centering
\small
\begin{tabular}{ll|cccccccc|l@{\hspace{5pt}}c}

\hline
 \textbf{Setup} & \textbf{Model} & \textbf{REL} & \textbf{ORG} & \textbf{VOC} & \textbf{STY} & \textbf{DEV} & \textbf{MEC} & \textbf{GRA} & \textbf{HOL} & \textbf{Avg$^{-H}$}&\textbf{Avg}  \\
 \hline
 \multirow{3}{*}{\textbf{Zero-shot}} & ALLaM & \textbf{0.242} & 0.197 & 0.319 & 0.309 & 0.335 & \uline{0.341} & \uline{0.312} & 0.365  &0.294& 0.303 \\
  & R7B & \textbf{0.242 }& \uline{0.348} & \uline{0.350} & \textbf{0.417} & \uline{0.340} & 0.150 & 0.292 & \uline{0.367} & \uline{0.306}& \uline{0.313} \\
  & Fanar & \uline{0.145} &\textbf{ 0.424} & \textbf{0.429} & \uline{0.388} & \textbf{0.411 }& \textbf{0.378} & \textbf{0.415} & \textbf{0.481} &\textbf{0.370} & \textbf{0.384}\\ 
\hline
\hline
 \multirow{9}{*}{\textbf{Prompt-specific}} & LR & 0.436 & 0.694 & 0.722 & {0.710} & 0.650 & 0.661 & {0.687} & 0.765 & 0.651 & 0.665 \\
  & RF & 0.519 & 0.704 & {0.734} & 0.718 & 0.670 & 0.658 & 0.694 & {0.775}  &0.671& 0.684\\
  & XGB & 0.522 & {0.716} & 0.743 & 0.731 & 0.685 & 0.686 & 0.711 & 0.791&0.685 & {0.698} \\
  & NN & 0.500 & 0.723 & 0.758 & 0.743 & 0.699 & 0.699 & \uline{0.727} & \uline{0.797} &0.693& 0.706 \\ \cline{2-12} 
  & AraT5 & \textbf{0.612} & \uline{0.731} & \uline{0.759} & \uline{0.759} & \uline{0.717} & \uline{0.709} & 0.719 & 0.787 &\uline{0.715} & \uline{0.724}\\
  & AraBERT & \uline{0.587} & \textbf{0.761} &\textbf{0.776}	&\textbf{0.768} &\textbf{0.731} &\textbf{0.724} &\textbf{0.744}	&\textbf{0.833}  &\textbf{0.727}&\textbf{0.740}\\ \cline{2-12} 
  & ALLaM (5) & 0.260 & 0.391 & 0.348 & 0.415 & 0.386 & 0.355 & 0.369 & 0.439&0.361  & 0.370\\
  & R7B (5) & 0.337 & 0.472 & 0.383 & 0.482 & 0.399 & 0.407 & 0.431 & 0.496 &0.416& 0.426 \\
  & Fanar (5) & {0.490} & 0.559 & 0.562 & 0.614 & {0.575} & {0.572} & 0.580 & 0.644 &0.565 & 0.575\\ 
\hline
\hline
 \multirow{10}{*}{\textbf{Cross-prompt}} 
  & LR & {0.360} & {0.621} & 0.633 & \uline{0.643} & 0.573 & {0.585} & {0.616} & 0.661  &0.576& {0.586} \\
  & RF & 0.331 & 0.609 & \textbf{0.644} & 0.637 & 0.573 & 0.559 & 0.609 & \textbf{0.682} &0.566& 0.581  \\
  & XGB & {0.360} & \textbf{0.645} & {0.641} & {0.641} & {0.583} & 0.577 & \uline{0.619} & \uline{0.679} &\uline{0.581}& \uline{0.593} \\
  & NN & 0.353 & 0.609 & 0.621 & 0.631 & 0.566 & 0.565 & {0.597} & {0.651}  &0.563& 0.574\\ \cline{2-12} 
  & ProTACT & {0.355} & {0.493} & {0.505} & {0.485} & {0.501} & {0.476} & 0.498 & 0.578&0.473 & {0.486} \\
  & AraBERT & 0.322 & 0.596 & 0.604 & 0.593 & 0.529 & 0.546 & 0.585 & 0.620 &0.539& 0.549\\ 
  & MOOSE & \textbf{0.411} & \uline{0.627} & \uline{0.642} &\textbf{0.649} & \uline{0.585} & \uline{0.586} & \textbf{0.623 }& 0.649  & \textbf{0.589}& \textbf{0.597}\\ \cline{2-12} 
  & ALLaM (5) & 0.131 & 0.237 & 0.213 & 0.287 & 0.226 & 0.249 & 0.251 & 0.281 &0.228 & 0.234\\
  & R7B (5) & 0.263 & 0.487 & 0.391 & 0.481 & 0.424 & 0.425 & 0.442 & 0.517 &0.416& 0.429 \\
  & Fanar (5) & \uline{0.409} & 0.580 & 0.541 & 0.623 & \textbf{0.588} & \textbf{0.592} & 0.587 & 0.669  &0.560& 0.574\\ 
\hline
  
\end{tabular}
\caption{Average QWK performance over all prompts for each trait under the zero-shot, prompt-specific, and cross-prompt setups. Avg$^{-H}$ denotes average performance without HOL scoring. \textbf{Bold} and \uline{underlined} values indicate the best and second-best performance per trait, respectively, marked separately within each setup. }
\label{tab:results}
\end{table*}

\paragraph{Cross-prompt Models}
We selected the same FB models and LLMs. For encoder-based models, we employ the same AraBERT-based architecture, which performed strongly in Arabic AES \cite{sayed2025feature}, along with two SOTA English AES models, ProTACT \cite{do-etal-2023-prompt} and MOOSE~\cite{chen-etal-2025-mixture-ordered}.


\paragraph{Feature Set}
We implemented the 816 handcrafted features introduced by \citet{sayed2025feature} for Arabic AES, encompassing surface, readability, lexical, semantic, and syntactic aspects. These features were applied to the FB models, ProTACT, MOOSE, and AraBERT. To mitigate noise, we performed feature selection based on Pearson and Spearman correlations \cite{li-ng-2024-conundrums}, retaining features whose absolute correlation with any trait exceeded a predefined threshold for either metric. For MOOSE, however, we did not apply this explicit feature selection step.


\paragraph{Evaluation \& Hyperparameters}
We evaluate model performance using QWK to measure agreement between human and model scores. Hyperparameters are tuned via Bayesian optimization with the Tree-structured Parzen Estimator algorithm \cite{NIPS2011_86e8f7ab}, implemented using Optuna's TPESampler.\footnote{\href{https://optuna.readthedocs.io/en/stable/reference/samplers/generated/optuna.samplers.TPESampler.html}{Optuna TPESampler documentation}} We ran 20 trials with 5 startup trials and a fixed random seed of 11. The best configuration, selected based on average QWK on the development set, was used for final evaluation on the test data. Additional hyperparameter details are provided in Appendix~\ref{appendix:baselines}.

\subsection{Results and Discussions} 
\label{sec:results-discussions}
We discuss the results of our benchmarking experiments for the two AES setups.

\subsubsection{Prompt-specific Results} 
\label{sec:prompt-specific-results}

\paragraph{Performance Across Categories (RQ1)} 
Table \ref{tab:results} summarizes performance differences across model categories. Among FB models, more complex architectures performed better, with NN leading, followed by XGB, RF, and LR. Both encoder-based models, AraT5 and AraBERT, outperform all other baselines, with AraBERT achieving SOTA performance (average QWK: 0.74) and AraT5 trailing by 1.6 points, highlighting their effectiveness in capturing prompt-specific patterns. For LLMs, adding prompt-specific few-shot examples improved the performance acorss all models. However, they still lagged behind, with the best variant, Fanar (5), scoring 16.5 points below AraBERT. 

\paragraph{Performance Across Traits (RQ2)} Encoder-based models consistently outperform other categories across traits, with AraBERT achieving the highest scores in 7 of 8 traits and AraT5 leading in REL. Their advantage over FB models is most pronounced in REL, where AraT5 and AraBERT exceed the best FB model by 9 and 6.5 points, respectively, demonstrating strength in modeling contextual and semantic relationships within essays.





\subsubsection{Cross-prompt Results} 
\label{sec:cross-prompt-results}

\paragraph{Performance Across Categories (RQ1)} FB models showed comparable performance, with only a 2-point difference between the top performer (XGB) and the lowest (NN), highlighting consistency of these feature-based approaches. LR's strong results further highlight the robustness and the quality of the feature set. Among encoder-based models, ProTACT exhibited the lowest performance overall, suggesting limited transferability from its English architecture, while AraBERT trailed XGB by 4.4 points. Notably, MOOSE achieved the highest performance among all evaluated cross-prompt models. For LLMs, few-shot prompting improved Fanar and R7B (+19 and +11.6 points), whereas ALLaM dropped by 7 points, indicating high sensitivity to prompt design.





\paragraph{Performance Across Traits (RQ2)} No single model consistently outperformed the others across individual traits. MOOSE performed best in REL, STY, and GRA traits. 
For the remaining traits, RF led in VOC and HOL, XGB in ORG, and Fanar (5) in DEV and MEC.

\subsubsection{Discussion}
The performance of models differ significantly between prompt-specific and cross-prompt setups. \textit{Encoder-based models} excel in prompt-specific tasks, approaching the IAA (Table \ref{tab:IAA}), which indicates a close reach to the human performance. For instance, AraBERT exhibits the best performance overall in the prompt-specific setup with an average QWK performance of 0.74. However, its cross-prompt performance drops significantly (average QWK: 0.549), underscoring its ability to capture prompt-dependent patterns rather than generalizable features. We note that the applicability of the prompt-specific setup is relatively limited due to reliance on target prompt essays that are labeled, which is often unavailable in practice. In the cross-prompt setup, while MOOSE achieves the highest average performance (average QWK: 0.597), we observe that classical FB models (XGB) remain remarkably competitive. This confirms that handcrafted features capture prompt-independent linguistic properties that generalize better in the more challenging and realistic cross-prompt setup than some standard encoders. 
LLMs, specifically Fanar and R7B with few shots, exhibit consistent performance across both setups, indicating they capture general scoring criteria. Their relative underperformance in prompt-specific tasks reflects their limited ability to exploit prompt-dependent patterns without fine-tuning. 

Overall, the findings of the benchmarking experiments above highlight the need to prioritize future research on cross-prompt AES, which is practically closer to the real world, while best models are still far from human performance.

\section{Conclusion}
\label{sec:conc}

While research on automated scoring of English essays began more than 55 years ago, Arabic essay scoring is hindered by the lack of data resources. To bridge this gap, this paper introduced \ds, the first large-scale publicly available dataset for Arabic AES. The dataset comprises 7,859 essays written on 8 different prompts by 4,372 high school students, and provides annotations for 7 writing traits as well as a holistic score with \textit{substantial} inter-annotator agreement. 
This makes it a \emph{comprehensive} and \textit{reliable} resource for training models and evaluating writing quality of Arabic essays. 
For reproducibility, we detailed the data collection and annotation process. We also benchmarked \ds{} using SOTA Arabic and English AES models in both prompt-specific and cross-prompt settings, showing strong baselines for future research. 

\section*{Acknowledgments}
We heartily thank our dedicated annotators for their contributions and express our gratitude to Qatar University Testing Center, the Ministry of Education and Higher Education in Qatar (the Arabic Section of the Department Of Educational Supervision in particular), participating schools, and students for making this work possible. We also acknowledge the support of Advanced Group for Information Technology (AGI) for providing access to the platform, Assessment Gourmet, which was used to manage and administer the annotation process. This work was made possible by NPRP grant NPRP14S-0402-210127 from the Qatar Research Development and Innovation (QRDI) Council. The statements made herein are solely the responsibility of the authors.

\section*{Limitations}
While \ds{} represents a significant step forward for Arabic AES, it has notable limitations. First, \ds{} was collected from students in a single country, Qatar, and specific grade levels, grades 10 to 12, which may limit its generalizability to other Arabic-speaking populations due to diverse educational systems. However, this concern is partially mitigated by Qatar's large population of Arab expatriates, resulting in student participants representing a wide variety of Arabic dialects and backgrounds. Second, \ds{} includes only explanatory and persuasive writing prompts, limiting genre diversity and potentially affecting model robustness across other styles, such as narrative or descriptive writing. Third, two of the prompts (Prompts 3 and 4) have fewer essays than the rest, which could introduce imbalance and lead to skewed performance in prompt-specific evaluations. Fourth, the relevance trait is scored on a narrow scale from 0 to 2, which limits the granularity of evaluation and may cause QWK to underrepresent subtle differences in model predictions. This restricted range likely contributes to the lower performance scores observed for the REL trait compared to others. Fifth, with an average essay length of only 171 words, the dataset lacks extended writing samples, restricting the assessment of models on long-form academic tasks requiring more complex, extended compositions. Finally, we acknowledge that the experimental results reported in Table \ref{tab:results} do not aim to exhaustively benchmark state-of-the-art AES models; rather, they serve as baseline performance for \ds{}. We emphasize that the primary contribution of this work lies in the construction and public release of a large-scale Arabic AES dataset, rather than in proposing novel AES model architectures.

\section*{Ethical Considerations}
The collection of \ds{} received approval from an Institutional Review Board (IRB Number: QU-IRB 159/2024-EA) and was conducted with informed consent from both students and their legal guardians. Essays were written under exam-style supervised conditions, with participation entirely voluntary and independent of students' academic evaluations or grades.
To protect participant privacy, all submissions were assigned pre-generated anonymous identifiers, and personally identifiable information was systematically removed during data cleaning. Annotation was conducted on a platform that prevented annotators from accessing student identities or demographic information. Essay prompts were carefully designed to be developmentally appropriate for the target age groups, culturally relevant to the Arabic-speaking context, and free of sensitive topics or potentially harmful content.
The dataset includes balanced gender representation and covers grades 10-12. 

All annotators were qualified Arabic language educators who received fair compensation at or above local professional rates. They completed structured training on annotation guidelines, and inter-annotator disagreements were resolved through systematic adjudication procedures.
The dataset will be released exclusively for research and educational purposes in Arabic automated essay scoring. We explicitly prohibit its use for high-stakes educational decisions affecting individual students. Users must commit to: (1) preventing re-identification attempts, (2) avoiding demographic profiling or inference, and (3) refraining from deployment in operational assessment contexts without independent validation. We recommend that any system developed using \ds{} undergo fairness auditing and stakeholder consultation before real-world application.

\bibliography{custom}
\bibliographystyle{acl_natbib}

\appendix
\section{Grading Rubric}
\label{app:rubrics}
For annotating \ds{}, we adopted the rubric from the Core Academic Skills Test (CAST), which is provided in Arabic,  developed by the Qatar University Testing Center (QUTC).\footnote{\url{https://www.qu.edu.qa/sites/en_US/testing-center/TestDevelopment/cast}} This rubric guided scoring across 7 writing traits: relevance (REL), organization (ORG), vocabulary (VOC), style (STY), development (DEV), mechanics (MEC), and grammar (GRA). An English-translated version of the CAST rubric is presented in Table \ref{tab:TansRubric}.

\begin{table*}[htbp]
\footnotesize
\centering
\begin{tabularx}{\linewidth}{|p{1cm}|X|X|X|X|X|}
\hline
\textbf{Trait} & \textbf{1} & \textbf{2} & \textbf{3} & \textbf{4} & \textbf{5} \\
\hline
\textbf{REL {\small \<الصلة\\ بالموضوع>}} & Partially relevant to the topic
& Completely relevant to the topic
& & & \\
\hline
\textbf{ORG  {\small\<الهيكل \\العام>}} & The introduction and conclusion are absent. There is no organization or sequence between paragraphs.
& Either the introduction or conclusion is absent. There is no organization or sequence between paragraphs.
& The text is well-organized and contains an introduction and conclusion, but the body has one paragraph (or two paragraphs) that lacks good coherence.
& The text is well-organized, contains an appropriate introduction and conclusion, and has two to three body paragraphs that are sequential and coherent.
& The text is well-organized and contains an introduction that introduces the topic, a conclusion that effectively concludes the text, and two to three body paragraphs that are sequential and well-connected. \\
\hline
\textbf{VOC {\small\<المفردات>}} & Use of a limited range of vocabulary and phrases that do not make sense together, with repetition and lexical errors, and generally inappropriate vocabulary that obscures meaning.
& Use of a basic range of vocabulary, with repetition, lexical errors, and many inappropriate choices that may obscure meaning.
& Use a sufficient range of vocabulary, with some repetition and lexical errors, with a small number of inappropriate vocabulary that may obscure meaning.
& Use of a good and appropriate range of vocabulary with few lexical errors, inappropriate choices without affecting meaning, and occasional use of idiomatic expressions.
& Use of a broad, correct, and appropriate range of vocabulary with few occasional errors, showing good knowledge of idiomatic expressions and awareness of implicit levels of meaning. \\
\hline
\textbf{STY {\small\<الأسلوب\\والتماسك\\البنائي>}} & The text employs very basic linear connecting words such as "and" and "then."
& Discourse develops as a simple list of points using only the most common connections.
& Discourse develops directly as a linear sequence of points using common structural cohesion devices.
& Discourse is clearly developed with main points supported by relevant details, appropriate use of different organizational patterns, and a range of structural cohesion devices.
& Discourse is well developed, with good inclusion of subtopics and details and a good conclusion, always appropriate use of a variety of organizational patterns, and a wide range of structural cohesion devices. \\
\hline
\textbf{DEV {\small\<تطور\\الأفكار\\والمضمون>}} & Content is not related to the subject; ideas are random and lack coherence, sequence, and evidence.
& Content is somewhat related; ideas are sequential, but the main idea disappears during writing, limited coverage, and poor use of supporting structures.
& Content is completely related; ideas mostly follow sequence, the main idea gradually disappears, some evidence is present but disorganized.
& Content is completely related; ideas are clear, organized, coherent, with the main idea connected to sub-ideas, a specific position adopted, some arguments and evidence presented coherently.
& Content is completely related; ideas are clear, organized, coherent, the main idea connected to sub-ideas, a specific position adopted, arguments and evidence presented coherently, comprehensive coverage of opinions, and use of various persuasive methods. \\
\hline
\textbf{MEC {\small\<الإملاء\\والترقيم\\والتنسيق>}} & Limited application of spelling rules.
& Frequent spelling and punctuation errors.
& Effectively applies standard formatting, paragraphing, spelling, and punctuation most of the time.
& Effectively applies standard formatting, paragraphing, spelling, and punctuation with few errors.
& Completely accurate paragraph organization, punctuation, and spelling, except for a few occasional pen slips. \\
\hline
\textbf{GRA {\small\<البناء\\والتراكيب>}} & Use a limited set of simple grammatical structures and sentence patterns with little flexibility or precision.
& Correct use of some simple structures with frequent systematic errors that may obscure meaning.
& Use a variety of grammatical structures, with notable errors that can sometimes obscure meaning.
& Good use of variety of structures with rare errors and minor imperfections that do not affect meaning.
& Always correct and flexible use of a wide variety of grammatical constructions with occasional minor slips. \\
\hline
\multicolumn{6}{|>{\hsize=\dimexpr5.5\hsize+5\tabcolsep\relax}X|}{\textbf{Note:} A score of 0 is assigned if the student does not attempt the task, provides a response that falls below the performance level described for score 1, or submits content that is not relevant to the topic of the given prompt. 
} \\

\hline
\end{tabularx}
\caption{CAST Persuasive/Argumentative Writing Rubric - English Translation \cite{bashendy-etal-2024-qaes}.}
\label{tab:TansRubric}
\end{table*}
 \begin{figure*}[t!]
    \centering
    \includegraphics[width=\textwidth]{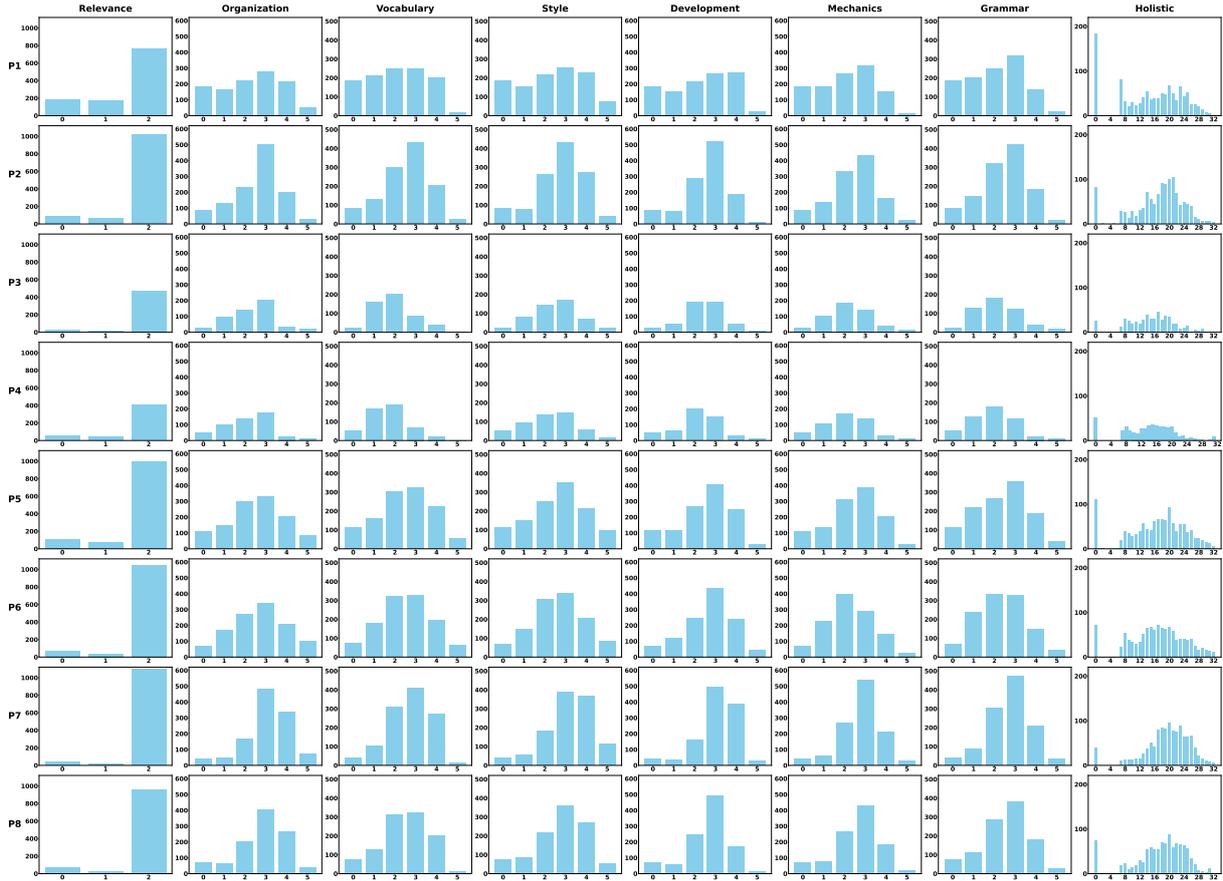}
    \caption{Score distributions across prompts and traits in \ds.}
    \label{fig:app-traits-dist}
\end{figure*}
\section{Detailed Dataset Analysis}
In this section, we present a detailed analysis of score distributions and essay lengths in \ds{}.
\label{sec:app_dataset_analysis}

\subsection{Score Distributions}
Figure \ref{fig:app-traits-dist} presents the score distributions in \ds{} across 8 distinct prompts (P1–P8) and 7 writing traits, including Relevance (REL), Organization (ORG), Vocabulary (VOC), Style (STY), Development (DEV), Mechanics (MEC), and Grammar (GRA), alongside a Holistic (HOL) score representing the sum of all individual traits. All traits are rated on a scale of 0–5, except REL, which ranges from 0–2. Overall, most traits approximate a normal distribution with comparable patterns across prompts, suggesting that essays' quality remains relatively stable regardless of the prompt topic. However, variations emerge by prompt and trait.

For REL, distributions are consistently skewed toward score 2 (the highest) across all prompts, indicating that most students successfully addressed the assigned task and produced strongly relevant essays. However, P1 and P5 display relatively higher frequencies at score 0, compared to other prompts, suggesting potential ambiguity in prompt wording that may have limited clear interpretation. Notably, both P1 and P5 are explanatory prompts, which often allow greater flexibility in structure and content. This openness may have led to varied interpretations among students and inconsistencies in alignment with the intended prompt focus.

The distribution of ORG scores varies across prompts but consistently peaks at score 3. P2 exhibits a pronounced peak at 3, whereas P3 and P4 show a wider spread toward lower scores, suggesting a weaker organizational structure in these essays. P5 and P6 display nearly identical, perfectly bell-shaped distributions, while P7 and P8 are skewed toward higher scores, reflecting stronger organization. Similarly, the VOC and STY traits generally follow comparable patterns, clustering around scores 2–3 with a common peak at 3, indicating that these dimensions present moderate challenges for most students.

The DEV, MEC, and GRA traits exhibit broadly comparable distributions across all prompts, though with subtle variations. The DEV trait shows distinct peaks at scores 2, 3, and 4, suggesting that the depth and elaboration of ideas vary depending on the prompt. In contrast, the MEC and GRA traits display closely aligned distributions, mirroring each other across prompts, with their peaks typically occurring at either score 2 or 3. This close alignment suggests a strong association between grammatical accuracy and mechanical correctness, as essays with better grammar tend to exhibit stronger adherence to writing conventions.

The HOL score, scaled up to 32, representing the sum of individual trait scores, provides a comprehensive measure of overall performance. Its distribution roughly approximates a bell curve, reflecting an overall normal pattern. However, prompt-specific effects are evident: P1, P3, P4, and P6 exhibit roughly centered distributions, whereas P2, P5, P7, and P8 are skewed toward higher scores, suggesting that students generally performed better on these prompts. This pattern indicates that certain prompts may have been easier or allowed for clearer demonstration of writing skills, while others elicited more variable performance. Also, it is worth noting that the scarcity of top scores (24+ holistically) suggests a ceiling effect, where achieving excellence demands strong performance across all dimensions.

Finally, the close uniform patterns across traits and prompts indicate a balanced scoring process, highlighting the robustness of the rubric and minimal influence from prompt variation or annotator bias. This consistency supports the development of generalized and robust Arabic AES models.

\subsection{Essay Length}
\begin{figure}[t]
    \centering
    \includegraphics[width=\columnwidth]{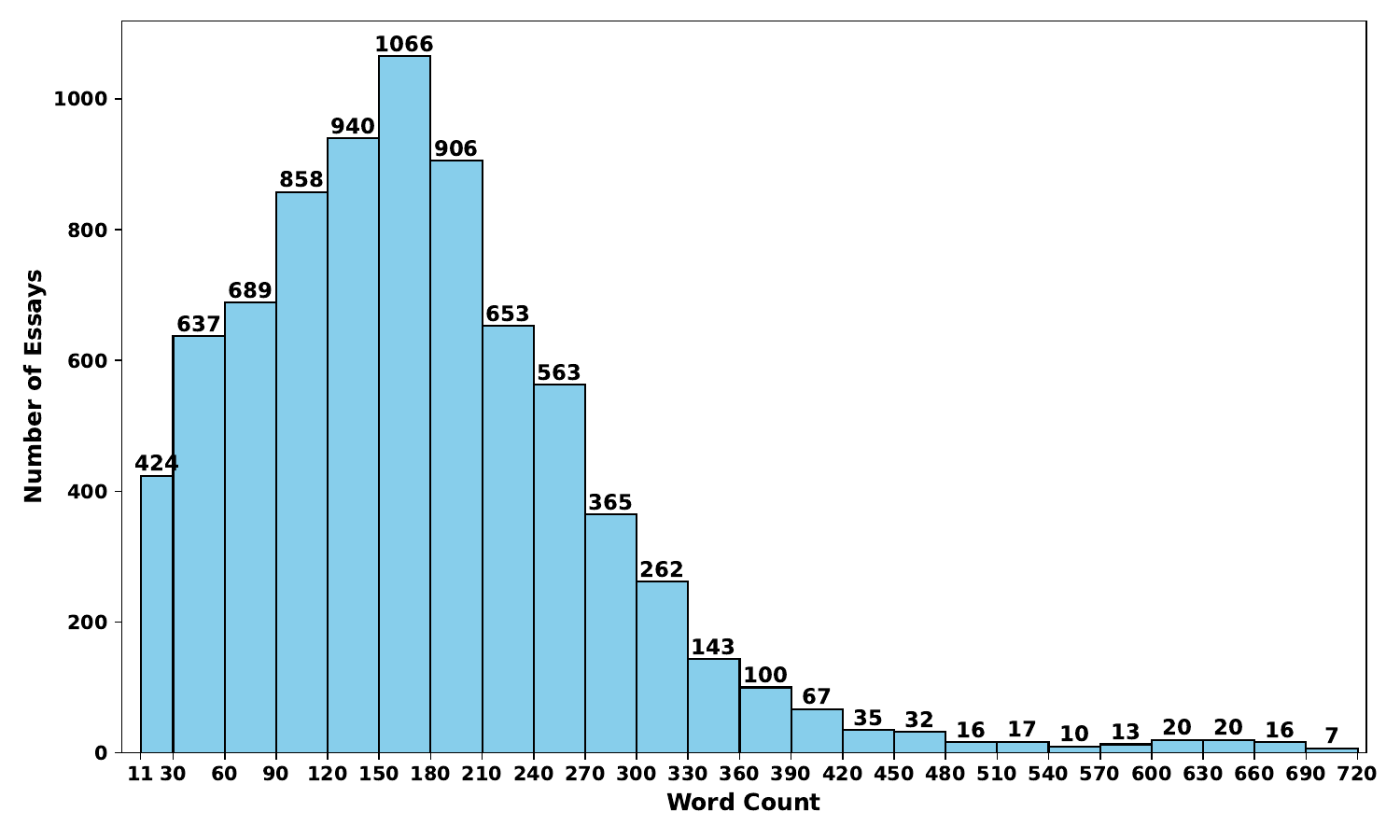}
    \caption{Distribution of essay lengths in \ds{}.}
    \label{fig:app-histogram}
\end{figure}

Figure \ref{fig:app-histogram} presents the distribution of essay lengths (in words) across all prompts in \ds{} dataset, which contains a total of 7,859 essays. The data exhibits a unimodal pattern centered around an average essay length of 171 words, with the majority of essays concentrated between 90 and 210 words. A pronounced peak occurs in the 150 to 180 word range, where the count reaches its highest at 1,065 essays. Shorter essays are also represented, with 1,061 essays falling between 11 and 60 words. Notably, the minimum observed length is 11 words, which results from the data cleaning step that excluded essays containing 10 words or fewer. Beyond 420 words, the frequency drops sharply, with only a few essays exceeding 500 words, suggesting that longer essays are less common in the dataset. Overall, this distribution provides a clear view of the typical essay length within \ds{} dataset, highlighting a strong central tendency and a sharp drop-off at the extremes.

\section{Implementation Details} \label{appendix:baselines}

In this section, we provide all the details of implementation and hyperparameter tuning for all the models used in the benchmarking experiments. Model-specific parameters are summarized in Table \ref{tab:hyperparameters}. All experiments were conducted on two machines: one equipped with an NVIDIA RTX A6000 GPU, and another with two NVIDIA A10 GPUs.


\subsection{Feature-based Models}
For the feature-based (FB) models, the architectures and hyperparameter configurations were kept consistent across both the \emph{prompt-specific} and \emph{cross-prompt} setups. 
We used the sklearn library\footnote{\url{https://scikit-learn.org/}} for Linear Regression (LR) and Random Forest (RF) models, and the XGBoost library\footnote{\url{https://xgboost.readthedocs.io/en/stable/}} for the Extreme Gradient Boosting (XGB) model. The Neural Network (NN) model followed the architecture described by \citet{li-ng-2024-conundrums}, consisting of two hidden layers with ReLU activations and a sigmoid output layer, implemented in PyTorch. For the NN, the number of epochs was fixed at 50 with early stopping, setting patience to 7, and checkpointing the best epoch during hyperparameter tuning for final model training. 

For all the FB models, feature selection was performed on the training set during hyperparameter tuning, with threshold values of [0, 0.1, 0.2, 0.3, 0.4, 0.5, 0.6], where 0 corresponds to using all features.

\begin{table*}[h!]
\centering
\small
            
\begin{tabular}{lll}
\hline
\textbf{Model} & \textbf{Hyperparameter Name} & \textbf{Value} \\
\hline
RF &Max depth &[3-10] with a step of 1\\
&Max features &[0.1-0.9] with a step of 0.1 \\
&Max samples &[0.1-0.9] with a step of 0.1 \\
\hline
XGB &Max depth &[3, 6, 9]\\
&Learning rate & [0.01, 0.1, 0.2] \\
&Subsample & [0.8, 0.9, 1.0] \\
&feature subsampling rate per tree &[0.8, 0.9, 1.0] \\
&Number of estimators & [100, 200, 300] \\
\hline
NN & Hidden layer widths & [64, 128, 256] \\
                & Dropout rate & 0.3 \\
                & Learning rate &[1e-5, 1e-4, 1e-3] \\
                & Batch size &16 \\
\hline
AraT5           & Learning rate & [5e-5, 1e-4, 2e-4] \\
                & Batch size  & 4 \\
               
\hline
AraBERT        
               & Batch size  & 8 \\
               & Encoder learning rate & [1e-6, 2e-6, 5e-6] \\
               & Dense-layers learning rate & [5e-5, 1e-4, 2e-4] \\
               & Number of trainable layers & [2, 4, 8, 'all'] \\
\hline
ProTACT        & Learning rate &[1e-5, 1e-4, 1e-3] \\
               & Trait similarity loss threshold &[0.5, 0.6, 0.7]\\
               & Batch size &16  \\
               & Alpha &0.7 \\
               & LSTM units & 32 \\
               & Self-attention heads & 4 \\
               & CNN filters & 100 \\
               & CNN kernel size & 3 \\
               & Dropout rate & 0.5 \\
\hline
MOOSE        & Learning rate &[1e-4, 2e-5] \\
              & Hidden dimensions & 256 \\
               & Batch size &8  \\
               & Weight decay & 0.001\\
               & Alpha &0.7 \\
               & Epochs & 14 \\
               & Chunk sizes & [10,30,90,130] \\
               & Dropout rate & 0.5 \\
\hline
\end{tabular}
\caption{Model-specific hyperparameters}
\label{tab:hyperparameters}
\end{table*}

\subsection{Encoder-based Models}

\subsubsection{AraT5}
Motivated by the strong performance of the T5 model in \emph{prompt-specific} English AES, we finetuned the AraT5 model using a similar setup to the ArTS model \cite{do-etal-2024-autoregressive}. 
The AES task is formulated as a text generation problem, where the model predicts the trait scores sequentially.
The input to the model consisted of an instruction to score the essay along with the essay text, and the model was fine-tuned to generate the trait names followed by their corresponding scores. 
The prediction order of the traits was aligned with the rubric to reflect the scoring order followed by human annotators.

Our setup differs from ArTS in data splitting: while their training and testing data included essays from all prompts (with a single model trained on all prompts), we adopted the conventional prompt-specific setup, training a separate model for each prompt using our predefined dataset splits.

For training, we used Seq2SeqTrainer\footnote{\url{https://huggingface.co/docs/transformers/en/main_classes/trainer##transformers.Seq2SeqTrainer}} from the transformers library.
For hyperparameter tuning, we explored the learning rates in the range [5e-5, 1e-4, 2e-4]. Other hyperparameters followed the configuration of \cite{do-etal-2024-autoregressive}, with a batch size of 4 and a maximum of 15 epochs. Early stopping was applied based on the average QWK score on the development set.

\subsubsection{AraBERT}
AraBERT is used as a baseline for the \emph{prompt-specific} and \emph{cross-prompt} setups using the same architecture and hyperparameter search space. The model is fine-tuned with a custom regression head and combined with handcrafted features for trait scoring.
The essay and prompt were provided as input to the encoder, with the max-pooled token representation concatenated with the handcrafted features. The resulting vector was then fed into eight parallel regression heads, each corresponding to a single trait score. Sigmoid activation is used at the output layer to produce values in the range [0, 1], which were subsequently rescaled to the appropriate range for each trait. 

For hyperparameter tuning, the learning rates for the encoder layers and the regression head are tuned separately to better accommodate the larger dataset, with a weight decay of 0.01. In addition, we tuned the number of trainable layers, with values of [2, 4, 8, all], where `all' corresponds to training all encoder layers. For the feature selection threshold, we used the same search space as that used with the FB models.  
These configurations were considered better suited to the dataset, mitigating training instabilities.

\subsubsection{ProTACT}
For ProTACT, we used the official implementation released by the authors.\footnote{\url{https://github.com/doheejin/ProTACT}} Essay representations were constructed using CNNs and LSTMs over Part-of-speech (POS) embeddings, while prompt representations combined POS and pre-trained GloVe embeddings. A multi-head attention mechanism was applied to obtain prompt-aware essay representations, which were then concatenated with handcrafted features and passed through a linear layer for scoring. The architecture was adapted for Arabic by replacing GloVe with AraVec embeddings~\cite{abubakr2017aravec}.\footnote{\url{https://github.com/bakrianoo/aravec}} POS embeddings were extracted using Camel Tools.\footnote{\url{https://camel-tools.readthedocs.io/}}

For hyperparameter tuning, we adopted the same search spaces for learning rate and feature selection threshold as in the NN model, while additionally tuning the trait similarity loss threshold. All other parameters were kept consistent with those reported in the original study, including embedding dimension, maximum essay length, maximum prompt length, LSTM units, self-attention heads, CNN filters, and kernel size.

\subsubsection{MOOSE}
For MOOSE, we used the official implementation released by the authors.\footnote{\url{https://github.com/antslabtw/MOOSE-AES}} The architecture was adapted for Arabic by replacing BERT with AraBERT \cite{antoun2020arabert}. Specifically, in the preprocessing stage, AraBERTv02\footnote{\url{https://huggingface.co/aubmindlab/bert-base-arabertv02}} was used only for tokenization and input encoding, while the finetuned model for the scoring task used AraBERTv01.\footnote{\url{https://huggingface.co/aubmindlab/bert-base-arabertv01}} This design choice was guided by empirical validation: we experimented with using AraBERTv01 and AraBERTv02 consistently for both preprocessing and modeling, as well as with mixed configurations, and found that using AraBERTv02 for tokenization and input encoding while finetuning AraBERTv01 for scoring yielded the best performance on the development set, thus it was adopted. For the incorporated features, We implemented the handcrafted features introduced by \citet{sayed2025feature} for Arabic AES, without applying any feature selection step. 

For hyperparameter tuning, the learning rate was tuned over the values 1e-4 and 2e-5, while all other parameters were kept consistent with those reported in the original study. Training was performed with a batch size of 8 for a maximum of 14 epochs, and no early stopping was applied. 

\definecolor{verylightgray}{gray}{0.98}

\begin{figure}[!t]
\centering
\small
\fcolorbox{black}{verylightgray}{
\begin{minipage}{\linewidth}
\tiny

\color{darkgray}
\begin{arabtex}
سيتم إعطاؤك مقال كتب ردًا على الموضوع المعطى. مهمتك هي تقييم جميع المعايير التالية للمقال.
\end{arabtex}

\vspace{0.3em}
\begin{flushright}
---
\end{flushright}

\color{magenta}
\begin{arabtex}
الموضوع: موضوع المقال

المقال: المقال المراد تقييمه.

الدرجات: الرجاء إعطاء الدرجات لجميع المعايير بهذا الشكل:
الصلة بالموضوع: <0-2>,
الهيكل العام: <0-5>,
المفردات: <0-5>,
الأسلوب والتماسك البنائي: <0-5>,
الأفكار والمضمون: <0-5>,
الإملاء والترقيم والشكل: <0-5>,
البناء والتراكيب: <0-5>
\end{arabtex}

\begin{flushright}
---
\end{flushright}

\color{olive}
\begin{arabtex}
\textbf{الموضوع}: هل تتفق أو تختلف جعلت الهواتف ورسائل البريد ...

\textbf{المقال}: إن مصطلح التكنولوجيا ...

\textbf{الدرجات}:
الهيكل العام: <3.0>,
المفردات: <3.0>,
الأسلوب والتماسك البنائي: <3.0>,
الأفكار والمضمون: <3.0>,
الإملاء والترقيم والشكل: <3.0>,
البناء والتراكيب: <3.0>,
الصلة بالموضوع: <2>
\end{arabtex}

\begin{flushright}
---
\end{flushright}

\begin{arabtex}
\textbf{الموضوع}: على الرغم من أهمية وسائل التواصل الاجتماعي ...

\textbf{المقال}: لا شك ان الافراط في استخدام وسائل التواصل ...

\textbf{الدرجات}:
الهيكل العام: <1.0>,
المفردات: <2.0>,
الأسلوب والتماسك البنائي: <2.0>,
الأفكار والمضمون: <2.0>,
الإملاء والترقيم والشكل: <1.0>,
البناء والتراكيب: <1.0>,
الصلة بالموضوع: <1>
\end{arabtex}

\begin{flushright}
---
\end{flushright}

\color{blue}
\begin{arabtex}
الموضوع: باتَ اِهْتمام وحماس المراهقين لِتعلُّم رِياضةٍ جديدة ...

المقال: الصحة والجسم السليم من نعم الله على الإنسان ...

الدرجات:
\end{arabtex}

\end{minipage}
}
\caption{An example of the LLM-prompt used in the few-shot setup, containing the \color{darkgray} base instructions, \color{black} the \color{magenta} input format, \color{black} the \color{olive} 5-shot examples, \color{black} and the \color{blue} input essay \color{black} for scoring. For the zero-shot setup, the  \color{olive} 5-shot examples \color{black} are removed, and all trait rubrics are included instead.}
\label{fig:llm_prompt}
\end{figure}

\subsection{LLMs}
The selection of the LLMs was based on the Open Arabic LLM Leaderboard,\footnote{\href{https://huggingface.co/spaces/OALL/Open-Arabic-LLM-Leaderboard}{\texttt{Open-Arabic-LLM-Leaderboard}}} where we selected the top 3 Arabic-centric open-weight models, at the time of the experiments, with a number of parameters < 10B. The selected models are: 
Fanar,\footnote{\href{https://huggingface.co/QCRI/Fanar-1-9B-Instruct}{\texttt{Fanar-1-9B-Instruct}}}
Command R7B Arabic,\footnote{\href{https://huggingface.co/CohereLabs/c4ai-command-r7b-arabic-02-2025}{\texttt{Command-R7b-Arabic}}} and 
ALLaM.\footnote{\href{https://huggingface.co/ALLaM-AI/ALLaM-7B-Instruct-preview}{\texttt{ALLaM-7B-Instruct-preview}}}

We evaluated the models under zero-shot and few-shot prompting. Figure \ref{fig:llm_prompt} illustrates the prompt adopted for our LLM scoring experiments. In the zero-shot setting, the LLM is tasked with generating trait scores in JSON format, given the prompt text, essay, and trait rubrics. The holistic score was computed as the sum of the individual trait scores. In the few-shot setting, the rubric was omitted, as the meaning of the scores could be inferred from the examples. The number of examples was fixed to 5, balancing scoring context with the context length limit (4096 tokens). When a prompt exceeded this limit, we iteratively removed examples until it fit. Although Command-R7B supports a larger context length, we fixed the number of examples to 5 to ensure fair comparison across LLMs.

In the prompt-specific setup, few-shot examples are selected from the same prompt. In the cross-prompt setup, examples are selected from different source prompts, ensuring that each comes from a distinct training prompt to expose the model to varied contexts. To account for variability in example selection, each experiment was repeated three times with random seeds 1, 11, and 42, and we report the average performance across runs.
For all the LLM experiments, we used vLLM\footnote{\url{https://docs.vllm.ai}} for inference with the outlines library\footnote{\url{https://dottxt-ai.github.io/outlines/latest/}} to enforce the JSON output format. 

\end{document}